# Creation of a Deep Convolutional Auto-Encoder in Caffe

Volodymyr Turchenko, Artur Luczak

Canadian Centre for Behavioural Neuroscience
Department of Neuroscience, University of Lethbridge
4401 University Drive, Lethbridge, AB, T1K 3M4, Canada
{vtu, luczak}@uleth.ca

***Abstract*** **-** The development of a deep (stacked) convolutional auto-encoder in the Caffe deep learning framework is presented in this paper. We describe simple principles which we used to create this model in Caffe. The proposed model of convolutional auto-encoder does not have pooling/unpooling layers yet. The results of our experimental research show comparable accuracy of dimensionality reduction in comparison with a classic auto-encoder on the example of MNIST dataset.

***Keywords*** – Deep convolutional auto-encoder, machine learning, neural networks, visualization, dimensionality reduction.

## 1. Introduction

The convolutional auto-encoder (CAE) is one of the most wanted architectures in deep learning research. As an auto-encoder, it is based on the encoder-decoder paradigm, where an input is first transformed into a typically lower-dimensional space (encoder part) and then expanded to reproduce the initial data (decoder part). It is trained in unsupervised fashion allowing it to extract generally useful features from unlabeled data, to detect and remove input redundancies and to present essential aspects of analyzing data in robust and discriminative representations [1]. Auto-encoders and unsupervised learning methods have been widely used in many scientific and industrial applications, solving mainly dimensionality reduction and unsupervised pre-training tasks. Compared to the architecture of a classic stacked auto-encoder [2], CAE may be better suited to image processing tasks because it fully utilizes the properties of convolutional neural networks, which have been proven to provide better results on noisy, shifted and corrupted image data [3]. Theoretical issues of CAE developments are well described in many research papers [1, 4-6].

Modern deep learning frameworks, i.e. ConvNet2 [7], Theano+Lasagne [8-9], Torch7 [10], Caffe [11] and others, have become very popular tools in the deep learning research community since they provide fast deployment of state-of-the-art deep learning models along with appropriate training strategies (Stochastic Gradient Descent, AdaDelta, etc.) allowing rapid research progress and emerging commercial applications. Our interest is to apply deep learning technologies, namely a CAE, for image processing in the neuroscience field. We have chosen Caffe deep learning framework mainly for two reasons: (i) a description of a deep neural network is pretty straightforward, it is just a text file with the description of layers and (ii) Caffe has a Matlab wrapper, which is very convenient and allows getting Caffe results directly into Matlab workspace for their further processing (visualization, etc) [12].

There are several existing solutions/attempts to develop and research a CAE model on different platforms, but to the best knowledge of the authors there is no current CAE implementation in Caffe yet. The issue of CAE implementation is permanently active in the Caffe user group [13-16]. There are two implementations of one-layer (not-stacked) CAE [17] and convolutional Restricted Boltzmann Machine [18] in Matlab. Mike Swarbrick Jones presented the implementation of CAE in Theano/Lasagne [19-20]. The Deep Sea team [21], who won a 100,000 US dollar prize (1[st] place) in the National Data Science Bowl, a data science competition where the goal was to classify images of plankton, has reported the use of CAE to pre-train only convolutional layers of their network (also implemented in Theano/Lasagne [9]). But they did not use this approach in their final winning architecture since they did not receive any substantial improvement with CAE pre-training. There are also CAE implementations: (i) in the examples of the Torch7 deep learning framework [22], (ii) recently implemented based on Torch7 [23], (iii) recently implemented on Theano/Keras [24] and (iv) in the examples of the Neon deep learning framework [25].

The goal of this paper is to present our first results of the practical implementation of a CAE model in Caffe deep learning framework as well as the experimental research of the proposed model on the example of MNIST dataset and a simple visualization technique which helped us to receive these results.

## 2. Creation CAE model in Caffe

In the examples of Caffe there are two models which solve the task of dimensionality reduction. The first is a classic stacked auto-encoder, proposed by Hinton et al [2], and the second is a Siamese network, proposed by LeCun et al [26]. The classic auto-encoder model is well researched and it trains in a purely unsupervised fashion. The Siamese network consists of two LeNet [3] architectures coupled in a Siamese way and ended by a contrastive loss function. Siamese network trains in a "semi-supervised" fashion, since for forming the training set we have to label a couple of input images (chosen randomly), which we are saving into two channels (left and right), by 1, if the images belong to the same class and by 0 otherwise. The visualizations how the example

Caffe implementations of the classic auto-encoder and the Siamese network solve the dimensionality reduction task encoding the test set (10000 examples) of MNIST in a 2-dimensional (2D) space are depicted in Fig. 1 and Fig. 2 respectively.

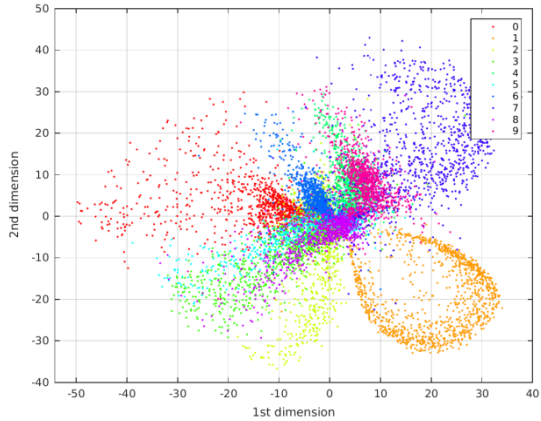

**Fig. 1. Visualization of MNIST test set in a 2D space by classic 2-dimensional auto-encoder**

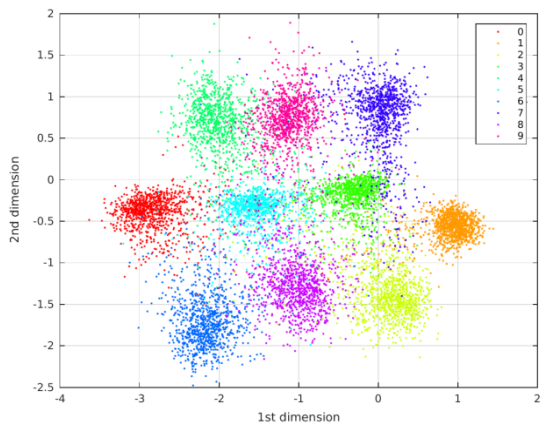

**Fig. 2. Visualization of MNIST test set in a 2D space by Siamese network**

Following the success of the LeNet [3] architecture and the paper [1] which showed experimentally that convolutional+pooling layers provide a better representation of convolutional filters and, furthermore, classification results, we started to construct CAE by the scheme *conv1-pool1-conv2-pool2* in the encoder part and *deconv2-unpool2-deconv1-unpool1* in the decoder part. We also have been inspired by the work of Hyeonwoo Noh et al [27] and we have used their non-official Caffe distribution [28], where they have implemented an unpooling layer, which is still absent in the official Caffe distribution. This CAE model did not train in our previous experiments (which are the topic of a separate paper) and, therefore, we have eliminated pooling-unpooling layers from our consideration. Masci et al [1] showed that convolutional architectures without max-pooling layers give worse results, but architectures without pooling and appropriate unpooling layers are definitely working architectures, and it is a good point to start first with some simpler working architecture and then to increase the complexity of the model.

After we have eliminated pooling-unpooling layers and added a non-linear activation function, *<Sigmoid>* in our case, after each convolutional and deconvolution layer [4], we have noticed, that the developed model is very similar to the classic auto-encoder model [2]. The difference is, the first two fully-connected layers of the encoder part have been replaced by two convolutional layers, and the last two fully-connected layers of the decoder part have been replaced by two deconvolution layers. The architecture of the developed CAE model in Caffe is depicted in Fig. 3. Taking into account some similarity between the classic auto-encoder and the developed model of CAE, we have used the following principles during our research:

1. The model should be symmetric in terms of the total size of feature maps and the number of neurons in all hidden layers in both the encoder and decoder parts. These sizes and numbers should decrease from layer to layer in the encoder part and increase in the same way in the decoder part similarly to a classic auto-encoder. These sizes and numbers should not be less than some minimal values allowing handling the size of the input data from the informational point of view;

2. Similarly to the example of the classic auto-encoder in Caffe, for the CAE model we have used two loss functions, *<Sigmoid_Cross_Entropy_Loss>* and *<Euclidean_Loss>*. Preliminary experimental research has shown that the use of only one of these loss functions separately does not provide good convergence results;

3. Visualization of the values (along with its numerical representation) of trainable filters, feature maps and hidden units from layer to layer allows better understanding of how data are converted/processed from layer to layer [29];

4. The main purpose of the activation function after each convolutional/deconvolution layer is non-linear data processing [4]. Since the nature of convolutional/deconvolution operations is a multiplication, our visualization showed the huge rise of a resulting value of convolutional/deconvolution operations (the values of feature maps) in encoder/decoder parts from layer to layer which prevents the CAE model from a desirable convergence during learning. So the use of activation functions, which drops the resulting value of feature maps to the interval [0...1] kept the values of feature maps at the end of the decoder part smaller in order to provide good convergence of the whole model.

5. The well-known fact is that good generalization properties of neural networks depend on the ratio of trainable parameters to the size and dimension of the input data. Therefore it is necessary to perform a set of experiments on the existing model of classic auto-encoder in order to find better architecture, (i.e. the size of trainable parameters and appropriate number of neurons in all hidden layers) which provides better generalization properties. Then, a CAE with similar size in terms of the total size of feature maps and the number of neurons in all hidden layers could be used to create the CAE model with good generalization properties. We cannot compare the classic auto-encoder and the CAE on the basis of the number of trainable parameters, because a convolutional network of the same size has much fewer trainable parameters [3];

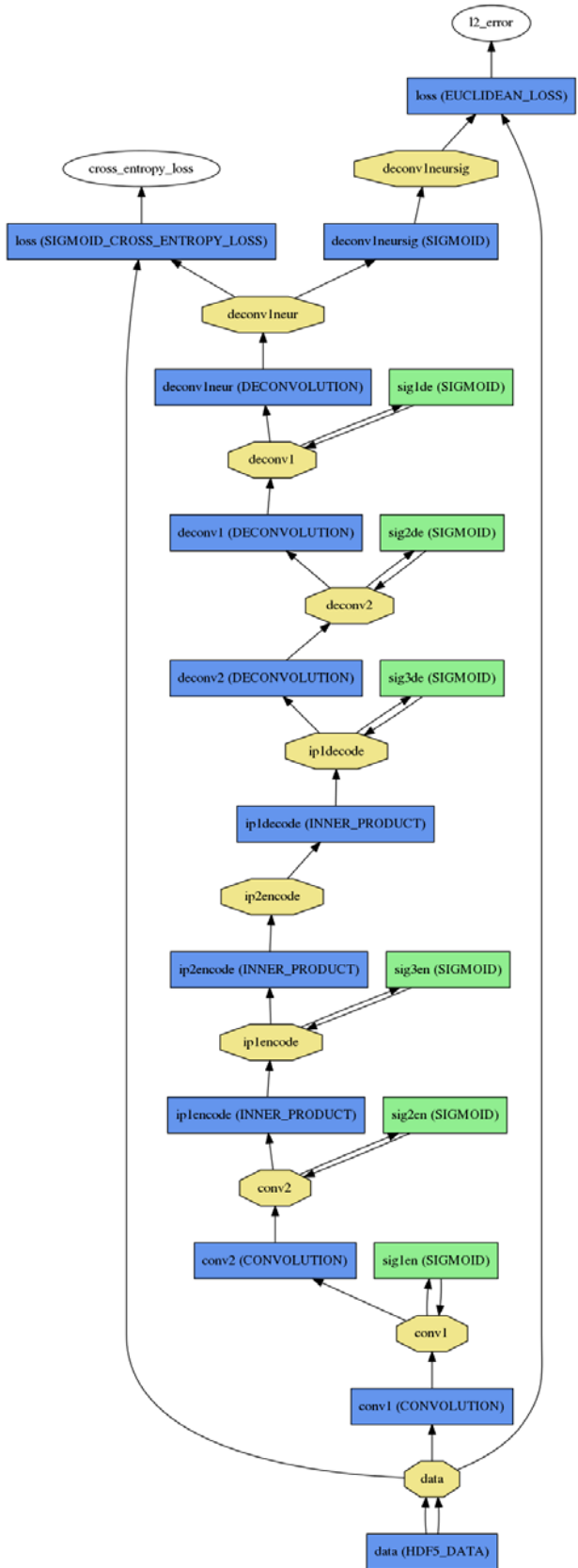

**Fig. 3. CAE model in Caffe**

6. The created architecture should be stable. Since different runs of a neural network may show different learning convergences (values of loss function) depending on random initialization of weights/biases, under stable architecture we mean the same convergence results within several (at least three) runs of the same model;

The practical implementation of these principles and some experimental results are presented in the next section.

## 3. Experimental results

All experimental researches were fulfilled on a workstation operated under Ubuntu 14.04.2 operation system. The workstation is equipped with 4-core (8 threads visible in Linux) Inter(R) Xeon(R) E5620@2.40 GHz processor, 51 Gb of RAM and GeForce GTS 450 GPU. The GPU has Fermi architecture, 192 CUDA cores, 1 Gb of RAM and computing capability 2.1. The version of Caffe distributed by Noh et al. [28] was used in the experiments. The training of all presented models was performed in GPU mode (*solver_mode: GPU*), thus one CPU core + GPU device were utilized by Caffe during the training.

The results of experimental tests of different sizes (further - architectures) of classic auto-encoder from Caffe examples are presented in Table 1. We were sure that Hinton et al. presented the best architecture (third row in Table 1) in their Science paper [2], but, as mentioned above, we wanted to research the generalization properties of smaller and bigger architectures in order to use this experience to create our CAE model. The number of trainable parameters for each architecture is specified in the first column. Here in Table 1 and in all next Tables, all calculations are provided for the case N = 2 dimensions. The size of the MNIST training dataset we calculated as 60000 examples x 784 elements = 47040 K elements. The ratio of MNIST size to the number of the training parameters (w+b) is specified in the second column. We trained each architecture in three fashions, with 2, 10 and 30 (30 is exactly as in Caffe examples) neurons in the last hidden layer of the encoder part, which corresponds to 2, 10-, 30-dimensional space of encoding respectively. According to Hinton et al [2] these architectures are called 2-, 10- and 30-dimensional auto-encoders. We did three runs for each architecture, the values of both loss functions, separated by ";" are presented in red for the training set and in black for the test set. In the last column we specified the number of classes we evaluated visually (see Fig. 1) from three runs of each architecture. For example, the number '25' means that 25 classes from 30 in total (10 classes per three runs) were correctly classified. Under 'correctly classified' we mean that the color of one class is not mixed with the color of another class(es), but this is, of course, a subjective sentiment.

We have used t-SNE technique [30] to visualize 10- and 30- dimensional data, produced by auto-encoder. For that visualization (it is showed later on the example of CAE in Fig. 5-6) we have integrated both calls, Caffe and t-SNE, into one Matlab routine.

**Table 1. Results of experimental research of classic auto-encoder from Caffe examples**

| Architecture, Number of trainable parameters (w+b), Auto-encoder size, elements | Data /(w+b) | LOSS values, (SIGMOID_CROSS_ENTROPY_LOSS; EUCLIDEAN_LOSS), train / test | | | | Visuali-zation, classes |
|---|---|---|---|---|---|---|
| | | N, Dims | Run 01 | Run 02 | Run 03 | |
| 784-300-150-75-N-75-150-300-784, 584254, 1052 | 81/1 | 2 | 150.67; 17.40 156.15; 18.12 | 150.45; 17.40 159.97; 18.85 | 148.21; 17.02 157.87; 18.15 | 14 |
| | | 10 | 110.50; 10.14 113.60; 10.56 | 108.04; 9.75 112.22; 10.15 | 113.24; 10.64 116.13; 11.00 | 22 |
| | | 30 | 107.03; 9.56 110.76; 10.03 | 106.44; 9.38 110.25; 9.92 | 104.26; 9.09 107.64; 9.51 | 26 |
| 784-500-250-125-N-125-250-500-784, 1098754, 1752 | 43/1 | 2 | 147.55; 16.75 152.79; 17.44 | 146.48; 16.70 155.13; 17.97 | 144.91; 16.43 153.05; 17.56 | 17 |
| | | 10 | 105.97; 9.39 109.39 ; 9.85 | 103.69; 8.99 106.32; 9.35 | 103.59; 9.05 106.45; 9.36 | 23 |
| | | 30 | 93.55; 7.34 95.66; 7.51 | 94.38; 7.45 97.18; 7.76 | 90.60; 6.82 93.06; 7.08 | 25 |
| 784-1000-500-250-N-250-500-1000-784, 2822504 3502 | 17/1 | 2 | 142.99; 16.08 152.02; 17.40 | 143.33; 16.15 152.68; 17.53 | 147.90; 16.85 151.33; 17.20 | 21 |
| | | 10 | 105.68; 9.36 109.05; 9.80 | 102.25; 8.85 105.16; 9.15 | 107.80; 9.74 111.04; 10.16 | 25 |
| | | 30 | 89.51; 6.64 91.75; 6.87 | 92.91; 7.23 95.75; 7.53 | 97.33; 7.94 99.82; 8.20 | 27 |
| 784-2000-1000-500-N-500-1000-2000-784, 8145004 7002 | 6/1 | 2 | 152.25; 17.54 161.82; 18.98 | 202.58; 26.01 205.81; 26.44 | 202.58; 26.01 205.80; 26.44 | 5 |
| | | 10 | 130.65; 13.62 136.55; 14.44 | 139.03; 15.33 148.95; 16.85 | 201.60; 25.83 204.91; 26.28 | 12 |
| | | 30 | 129.13; 13.39 134.96; 14.17 | 157.92; 18.44 166.55; 19.80 | 129.75; 13.54 135.23; 14.25 | 16 |
| 784-3000-1500-750-N-750-1500-3000-784, 15967504 10502 | 3/1 | 2 | 202.64; 26.03 205.80; 26.44 | 202.64; 26.03 205.80; 26.44 | 202.64; 26.03 205.80; 26.44 | 3 |
| | | 10 | 202.64; 26.03 205.81; 26.44 | 202.58; 26.03 205.75; 26.43 | 202.66; 26.03 205.82; 26.44 | 7 |
| | | 30 | 200.96; 25.70 204.28; 26.16 | 202.64; 26.03 205.80; 26.44 | 202.64; 26.03 205.80; 26.44 | 15 |

The learning parameters of the solver of the classic auto-encoder (in file *solver.prototxt*) we left as it is specified in Caffe examples. These parameters were: solver_type: SGD (by default), base_lr: 0.01, lr_policy: "step", gamma: 0.1, stepsize: 1000, momentum: 0.9 and weight_decay: 0.0005. The results in Table 1 are specified for 5000 training iterations. The training time for the architecture 784-1000-500-250-N-250-500-1000-784 from the third row of Table 1 was 2 minutes. As we can see from Table 1, the architectures with bigger ratio Data/(w+b), namely 81/1, 43/1 and 17/1 have better generalization properties because they provide lower (better) values of loss functions during the training and testing and better visualization results.

It is necessary to note, that in the case of the classic fully-connected auto-encoder researched above, the number of trainable parameters (except biases) is equal to the number of connections. But this is not a case of a CAE since convolutional/deconvolution layers have much fewer trainable parameters, because the sizes of convolutional and deconvolution kernels are the trainable parameters in case of CAE. Therefore, in order to build a CAE, we have operated with the term "CAE size" which is the total number of elements in feature maps and the number of neurons in all hidden layers in the encoder and decoder parts. In this paper we present two architectures: (i) Model 1, where we more-or-less adjusted the CAE size to the best architecture of the classic auto-encoder (third row in Table 1) and (ii) Model 2, where we increased the CAE size after the series of experimental tests of Model 1.

Table 2 contains the architecture parameters for both developed CAE models. In both models we did convolution with kernels *conv1* 9x9 and *conv2* 9x9 in the encoder part and we came back with deconvolution kernels *deconv2* 12x12 and *deconv1* 17x17 in the decoder part (Fig. 1). We have chosen this size of deconvolution kernels in order to restore the same size of

MNIST image 28x28. The only difference between the two models is the number of feature maps in *conv/deconv* layers and the number of neurons in fully-connected layers in the encoder and decoder parts. In the third column of Table 2 we can see (bold font) that both proposed models are practically symmetric in terms of the total number of elements in feature maps and the number of neurons in all hidden layers of the encoder and decoder parts.

The number of trainable parameters for both models and the details of their calculation are specified in the last column of Table 2 and in Table 3 respectively. Since the deconvolution operation has the same nature as convolution [31], we have used the same approach to calculate the number of trainable parameters both in the encoder and decoder parts. These calculations can be easily checked by calling Caffe from Matlab using the command *<caffe('weights');>*. In decoder part, the purpose of the deconvolution layer *deconv1neur*, which corresponds to the term (1*1w+1b) in the third column of Table 3, is to transform all feature maps of the last deconvolution layer *deconv1* into one restored image with the same size as the original: 28x28 pixels in case of MNIST. There was an explanation in the Caffe user group how to do that [32]. As we can see from Tables 2 and 3, the proposed CAE models are practically symmetric not only in terms of the total number of elements in feature maps and the number of neurons in the hidden layers, but also in terms of the number of trainable parameters in both the encoder and decoder parts. The comparison with a similarly-sized classic auto-encoder (3502, third row of Table 1) and Model 1 of the developed CAE (3996, Table 2) shows that the CAE has (2822504/74893 =) 38 times fewer trainable parameters.

**Table 2. Architecture parameters of two CAE models**

| CAE | Architecture | Size of feature maps and number of hidden nodes | CAE size, elements | Number of trainable parameters (w+b) |
|---|---|---|---|---|
| Model 1 | 784-(9x9x4)-(9x9x2)-125-N-125-(12x12x2)-(17x17x2)-784 | (20x20x4)-(12x12x2)-125-N-125-(12x12x2)-(28x28x2) | 3996 | 74893 |
| | | **1600-288-125-N-125-288-1568** | | |
| Model 2 | 784-(9x9x8)-(9x9x4)-250-N-250-(12x12x4)-(17x17x4)-784 | (20x20x8)-(12x12x4)-250-N-250-(12x12x4)-(28x28x4) | 7990 | 297391 |
| | | **3200-576-250-N-250-576-3136** | | |

**Table 3. Calculation of the number of trainable parameters in both encoder and decoder parts**

| CAE | Number of trainable parameters, w(weights)+b(biases), i(inputs), o(outputs) | | |
|---|---|---|---|
| | Encoder part | Decoder part | Total |
| Model 1 | *conv1*->((9*9w+1b)*1i*4o)+ *conv2*->((9*9w+1b)*4i*2o)+ *ip1encode*->(288i*125o+125b)+ *ip2encode*->(125i*2o+2b) = (324w+4b)+(648w+2b)+(36000w+125b) +(250w+2b) = **37355** | *ip1decode*->(2i*125o+125b)+ *deconv2*->((12*12w+1b)*125i*2o)+ *deconv1*->((17*17w+1b)*2i*2o)+ *deconv1neur*->((1*1w+1b)*2i*1o)+ (250w+125b)+(36000w+2b)+(1156w+2b)+ (2w+1b) = **37538** | ***74893*** |
| Model 2 | *conv1*->((9*9w+1b)*1i*8o)+ *conv2*->((9*9w+1b)*8i*4o)+ *ip1encode*->(576i*250o+250b)+ *ip2encode*->(250i*2o+2b) = (648w+8b)+(2592w+4b)+ (144000w+250b)+(500w+2b) = **148004** | *ip1decode*->(2i*250o+250b)+ *deconv2*->((12*12w+1b)*250i*4o)+ *deconv1*->((17*17w+1b)*4i*4o)+ *deconv1neur*->((1*1w+1b)*4i*1o)+ (500w+250b)+(144000w+4b)+(4624w+4b)+ (4w+1b) = ***149387*** | ***297391*** |

The results of experimental tests of the two developed CAE models are presented in Table 4, which is organized similarly to Table 1. We evaluate 2-, 10- and 30-dimensional CAEs, the number of dimensions corresponds to the number of neurons in the last hidden layer *ip2encode* (see Fig. 3) of the encoder part. The results in Table 4 are specified for 20000 training iterations. The comparison of both tables shows that Model 1 provides the same minimum values of loss functions as well as the same number of visualization classes as the best architecture of the classic auto-encoder (3[rd] row) from Table 1. The experimental tests of Model 2 showed better (lower) values of loss functions reached during the training, and slightly better visualization results in comparison with Model 1. The visualizations showing how Model 2 solves the dimensionality reduction task encoding the test set of MNIST in a 2D space are depicted in Figs. 4-6. Similarly we have used t-SNE technique [30] to visualize 10- and 30- dimensional data, produced by the 10- and 30-dimensional CAEs. In both cases, i.e. for the research of classical auto-encoder and the developed CAE, we have reformatted the original MNIST dataset into HDF5 format since this data format is perfectly supported by Matlab. The learning parameters of the solver of the developed CAE with stable architecture were: solver_type: SGD, base_lr: 0.006, lr_policy: "fixed", and weight_decay: 0.0005. We run several experiments, changing the architecture and learning parameters of CAE, but in many cases they were not stable architectures. For example, we tried different initializations of weights and biases (*<weight_filler>* and *<bias_filler>*). The presented results were provided with the following initialization: *<bias_filler {type: "constant"}>* for all layers, *<weight_filler {type: "xavier"}>* for convolutional/deconvolutional layers,

*<weight_filler {type: "gaussian" std: 1 sparse: 25}>* for fully-connected (InnerProduct) layers. Also we have tried *<ReLU>* activation functions instead of *<Sigmoid>* and, surprisingly, we have received worse results.

The accepted models and learning parameters are not unique, we are pretty sure there are a lot of other configurations, which could provide stable CAE architectures. The training times for the Model 1 and Model 2 running for 20000 training iterations were 68 and 100 minutes respectively. For quick reference we have collected all learning parameters of the classic auto-encoder and the developed CAE in Table 5.

**Table 4. Results of experimental research of developed CAE**

| Architecture, Number of trainable parameters (w+b), CAE size | LOSS values, (SIGMOID_CROSS_ENTROPY_LOSS; EUCLIDEAN_LOSS), train / test | | | | Visuali-zation, classes |
|---|---|---|---|---|---|
| | N, Dims | Run 01 | Run 02 | Run 03 | |
| Model 1, 784-(20x20x4)-(12x12x2)-125-N-125-(12x12x2)-(28x28x2)-784, 74893, 3996 | 2 | 152.68; 17.95 159.47; 18.71 | 151.83; 17.61 160.10; 18.76 | 156.65; 18.46 159.76; 18.66 | 20 |
| | 10 | 105.14; 9.33 109.58; 9.88 | 107.86; 9.72 111.80; 10.24 | 109.37; 9.91 113.43; 10.39 | 26 |
| | 30 | 94.53; 7.37 98.39; 7.78 | 96.02; 7.57 98.18; 7.76 | 93.58; 7.14 96.80; 7.53 | 26 |
| Model 2, 784-(20x20x8)-(12x12x4)-250-N-250-(12x12x4)-(28x28x4)-784, 297391, 7990 | 2 | 142.50; 15.93 148.74; 16.85 | 143.50; 16.20 148.33; 16.72 | 143.16; 16.17 149.68; 16.99 | 22 |
| | 10 | 90.72; 6.93 93.12; 7.15 | 91.34; 7.04 93.20; 7.19 | 93.27; 7.35 94.91; 7.42 | 26 |
| | 30 | 71.38; 3.82 72.52; 3.86 | 70.54; 3.66 71.65; 3.69 | 71.82; 3.88 73.65; 4.00 | 26 |

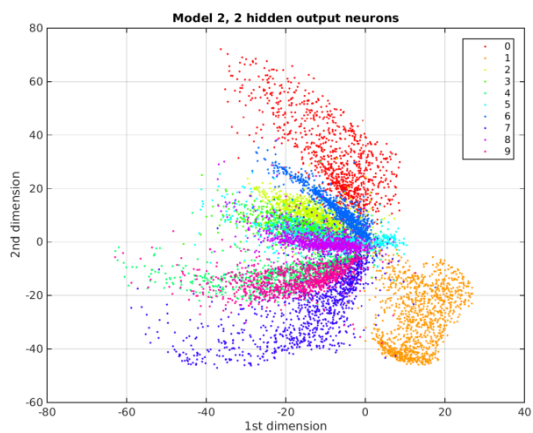

**Fig. 4. Visualization of MNIST test set in a 2D space by 2-dimensional CAE Model 2**

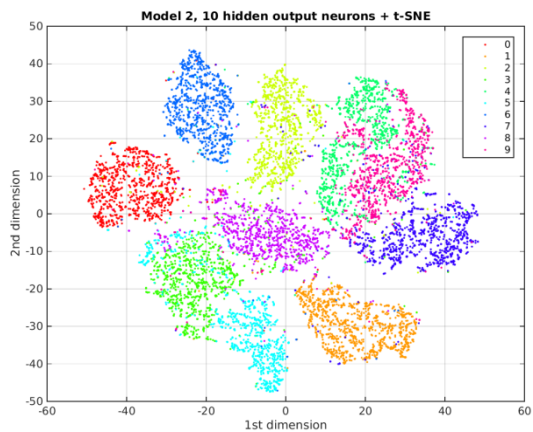

**Fig. 5. Visualization of MNIST test set in a 2D space by 10-dimensional CAE Model 2 + t-SNE**

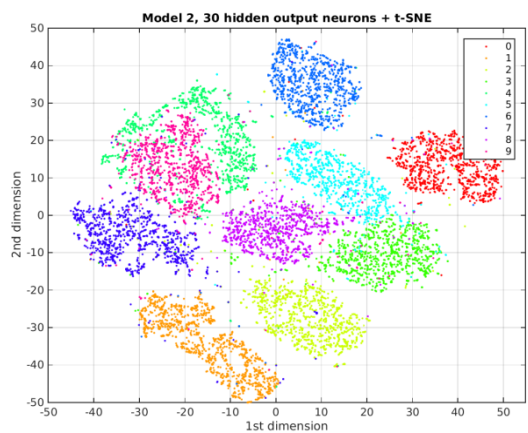

**Fig. 6. Visualization of MNIST test set in a 2D space by 30-dimensional CAE Model 2 + t-SNE**

As mentioned above, during experimental research we have visualized the feature maps and the outputs of the hidden layer neurons in order to understand how data are processing inside the CAE. The implementation of such visualization is simple and straightforward thanks to the Matlab wrapper in Caffe. We just created 15 *.prototxt* files corresponding to the number of CAE layers in Fig. 3. After training, and having an appropriate *.caffemodel* file, we call Caffe in Matlab using each of those *.prototxt* files as an argument. Received values, produced by each layer were visualized. An example showing how CAE Model 2 encodes and decodes the digit "2" is depicted in Fig. 7. The left upper picture is the original image with 28x28 pixels and the right bottom picture with the name *<Deconv1neursig>* is the restored image with 28x28 pixels. The title of each picture contains the following information: the digit we visualized (i.e. "2" - we have such pictures for 10 example digits from MNIST), the picture number (i.e. "01", "02", etc.), the size of the appropriate layer how it is internally represented in Caffe (i.e. "20x20x8"), the name of the appropriate layer corresponding to the names of the layers from Fig. 3 (i.e. "*Conv1*"). We also calculated the minimum and maximum values for each *conv/deconv* layer and specified them in square brackets in the titles of appropriate pictures. This allowed us to understand that in the failed experiments, the outputs of *deconv2* and *deconv1* layers were saturated, and therefore the pixels of the restored image had the value 0 and the loss values during training were NaN (Not A Number).

**Table 5. Learning parameters of classic auto-encoder and developed CAE in Caffe**

| | Learning parameters in file *solver.prototxt* | Training time | |
|---|---|---|---|
| Classic auto-encoder | base_lr: 0.01, lr_policy: "step", gamma: 0.1 stepsize: 1000, momentum: 0.9, weight_decay: 0.0005 | Architecture 784-1000-500-250-N-250-500-1000-784, 5000 training iterations | |
| | | **2 minutes** | |
| Developed CAE | base_lr: 0.006, lr_policy: "fixed", weight_decay: 0.0005 | Model 1, 20000 training iterations | Model 2, 20000 training iterations |
| | | **68 minutes** | **100 minutes** |

All appropriate *.prototxt* files of the developed CAE along with all Matlab scripts which we have used for all visualizations have been published in the Caffe user group [33] and Dr. A. Luczak's web-page [34]. It is necessary to note, that the developed CAE model is working on the version of Caffe used/distributed by Noh et al [24] (the date of the files in this version is Jun 16, 2015). We ran the CAE model on the latest version we have (the date of files in this version is Apr 05, 2016). It seems, in the newer versions after Jun 16, 2015, the Caffe developers have changed: (i) the syntax of layer descriptions – from "layers" to "layer", and layers' types from "CONVOLUTION" to "Convolution", etc. and (ii) the internal representation of fully-connected (InnerProduct) layers: it is a 2-dimensional array now, not 4-dimensional, as it was in the previous version(s). To deal with these issues it is necessary to change the syntax in the *.prototxt* files accordingly and to change the dimensionality of the last fully-connected layer before the first deconvolution layer in the decoder part using the *<reshape>* layer as follows: *<layer {name: "reshape" type: "Reshape" bottom: "ip1decode" top: "ip1decodesh" reshape_param { shape { dim: 0 dim: 0 dim: 1 dim: 1 }}}>*.

## 4. Conclusions

The development of a deep (stacked) convolutional auto-encoder in Caffe deep learning framework and its experimental evaluation are presented in this paper. The paper contains the first research results of our deep convolutional auto-encoder. The proposed model does not contain pooling/unpooling layers yet. In contrast to the classic stacked auto-encoder proposed by Hinton et al [2], convolutional auto-encoders allow using the desirable properties of convolutional neural networks for image data processing tasks while working within an unsupervised learning paradigm. The results of our experimental research show comparable accuracy in a dimensionality reduction task compared with the classic auto-encoder on the example of MNIST dataset.

During the creation of this convolutional auto-encoder we have used well-known principles, mentioned in Section 2 above, which are used by many machine learning researchers every day. Nevertheless, we believe that our approach and research results, presented in this paper, will help other researchers in general - and the Caffe user group in particular - to create efficient deep neural network architectures in future.

Application of the developed deep convolutional auto-encoder for our tasks in the neuroscience field and creation of more complex architectures with pooling/unpooling layers are the directions of our future research.

## Acknowledgements

We would like to thank the Caffe developers for creating such a powerful framework for deep machine learning research. We thank Hyeonwoo Noh and Dr. David Silver for discussions on some results presented in this paper, Dr. Eric Chalmers for the editorial help, and Dr. Robert Sutherland for help with financial support.

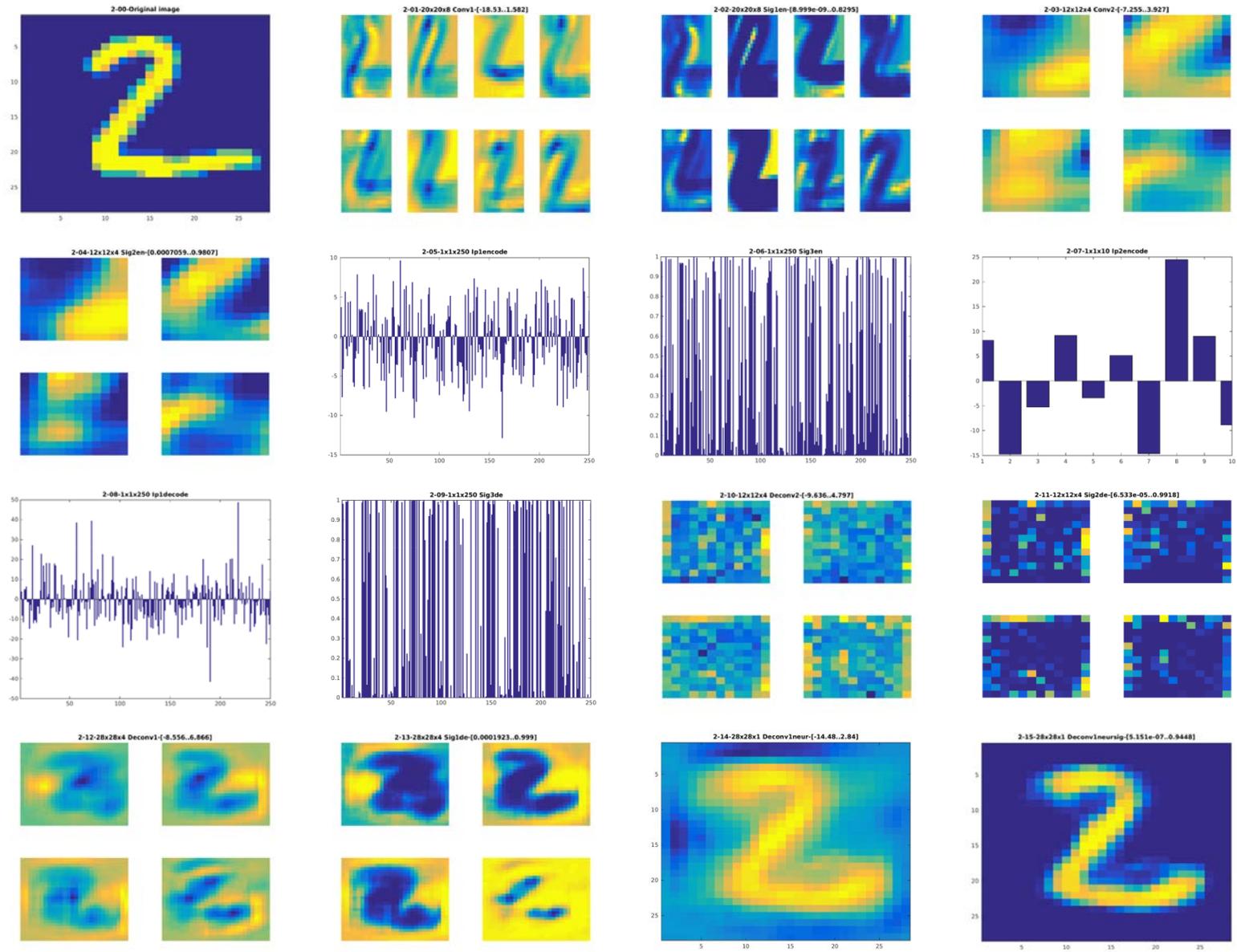

Fig. 7. Visualization of encoding and decoding digit "2" by 10-dimensional CAE Model 2

## Changes in this Version2 in comparison with Version1 published Dec 4, 2015

We changed/added references [23], [24], [25] about some CAE solutions that have appeared or we have founded after Dec 4, 2015. We added references [33] and [34] to specify the exact links where our *.prototxt* files were published. We corrected an inaccuracy in Table 3 in the calculation of the number of trainable parameters. We improved the text of several paragraphs through including better explanation of Fig. 7, and some tips on how to run the developed CAE in the newest version of Caffe.